\documentclass{article}
\usepackage{ijcai21}

\usepackage{times}
\usepackage{soul}
\usepackage{url}
\usepackage[hidelinks]{hyperref}
\usepackage[utf8]{inputenc}
\usepackage[small]{caption}
\usepackage{graphicx}
\usepackage{amsmath}
\usepackage{amsthm}
\usepackage{booktabs}
\usepackage{algorithm}
\usepackage{algorithmic}
\usepackage{multirow}
\usepackage{multicol}
\usepackage{mathrsfs}
\usepackage{amsfonts,amssymb}
\usepackage{subfigure}
\usepackage{color}
\usepackage{footnote}

\title{GM-MLIC: Graph Matching based Multi-Label Image Classification}

\author{
Yanan Wu$^1$\footnote{The authors contributed equally to this work.} \and
He Liu$^1$\footnotemark[1] \and
Songhe Feng$^1$\footnote{Corresponding Author}\and
Yi Jin$^1$\and
Gengyu Lyu$^1$\And
Zizhang Wu$^2$\\
\affiliations
$^1$Beijing Key Laboratory of Traffic Data Analysis and Mining \\
School of Computer and Information Technology, Beijing Jiaotong University\\
$^2$Zongmu Technology\\
\emails
\{19112034, liuhe1996, shfeng, yjin, 18112030\}@bjtu.edu.cn,
zizhang.wu@zongmutech.com
}

\begin{document}

\maketitle

\begin{abstract}

Multi-Label Image Classification (MLIC) aims to predict a set of labels that present in an image. The key to deal with such problem is to mine the associations between image contents and labels, and further obtain the correct assignments between images and their labels. In this paper, we treat each image as a bag of instances, and reformulate the task of MLIC as an instance-label matching selection problem. To model such problem, we propose a novel deep learning framework named Graph Matching based Multi-Label Image Classification (GM-MLIC), where Graph Matching (GM) scheme is introduced owing to its excellent capability of excavating the instance and label relationship. Specifically,
we first construct an instance spatial graph and a label semantic graph respectively, and then incorporate them into a constructed assignment graph by connecting each instance to all labels. Subsequently, the graph network block is adopted to aggregate and update all nodes and edges state on the assignment graph to form structured representations for each instance and label. Our network finally derives a prediction score for each instance-label correspondence and optimizes such correspondence with a weighted cross-entropy loss. Extensive experiments conducted on various image datasets demonstrate the superiority of our proposed method.
\end{abstract}

\section{Introduction}
Multi-label image classification (MLIC) is an essential computer vision task, aiming to assign multiple labels to one image based on its content. Compared with single-label image classification, MLIC is more general and practical since an arbitrary image is likely to contain multiple objects in the physical world. Thus, it widely exists in many applications such as image retrieval \cite{wei2019saliency} and medical diagnosis recognition \cite{ge2018chest}. However, it is also more challenging because of the rich semantic information and complex dependency of an image and its labels.

\begin{figure}[t]
\centering
\centerline{\includegraphics[scale=0.23]{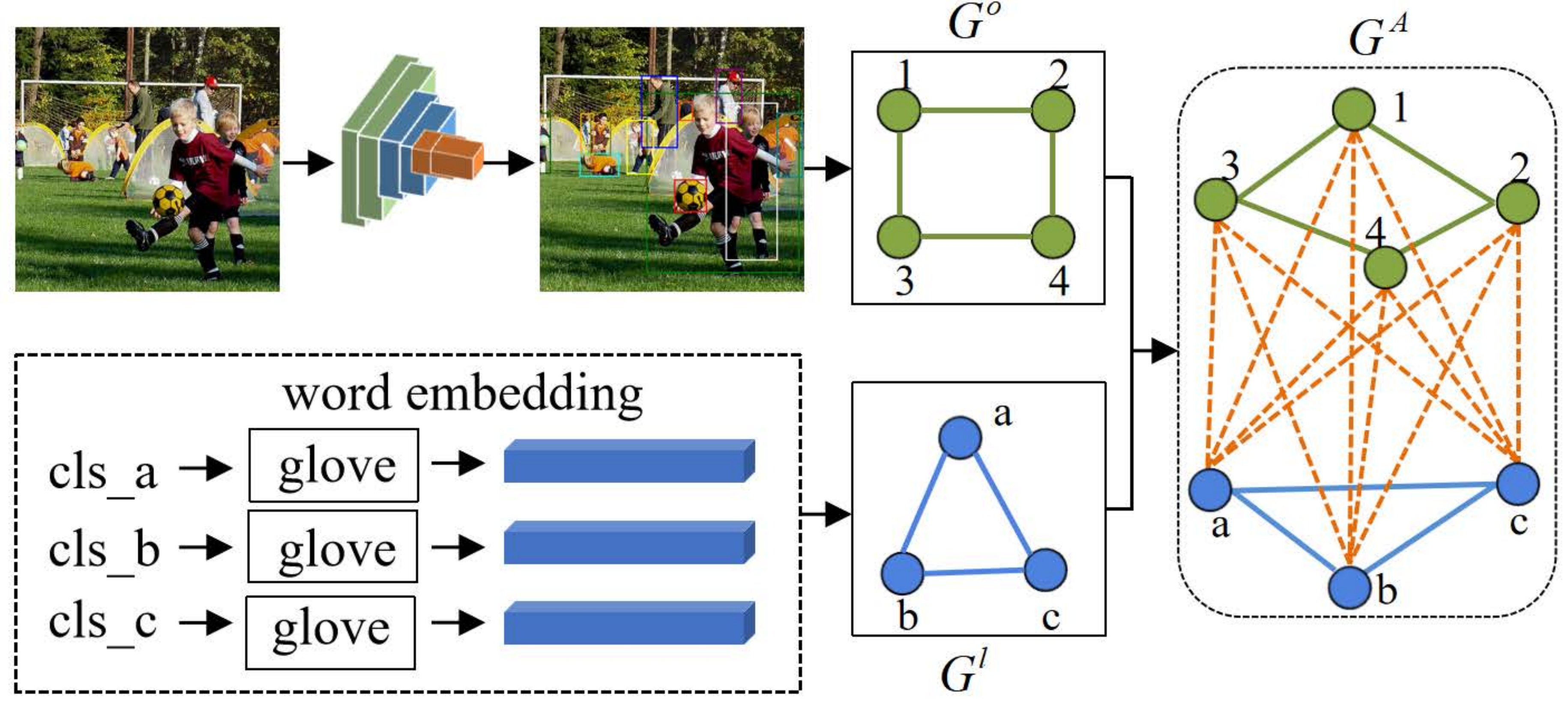}}
\vspace{-2mm}
\caption{Illustration of the ASsignment Graph Construction (ASGC). The upper part designs an instance spatial graph $\mathbb{G}^o$, while the lower part builds a label semantic graph $\mathbb{G}^l$. Then each instance is connected to all labels to form the final assignment graph $\mathbb{G}^A$.}
\label{ASGC}
\vspace{-5mm}
\end{figure}

The key to accomplish the task of MLIC is how to effectively explore the valuable semantic information from the image context, and further obtain the correct assignments between images and their labels. A simple and straightforward way is to treat each image as a bag of instances/proposals coping with the instances in isolation, and convert the multi-label problem into a set of binary classification problems. However, its performance is essentially limited due to ignoring the complex topology structure among labels. This stimulates research for approaches to capture and mine the label correlations in various ways. For example, Recurrent Neural Networks (RNNs) \cite{wang2016cnn} \cite{wang2017multi} and Graph Convolution Network (GCN) \cite{chen2019multi} \cite{wang2020multi} are widely used in many MLIC frameworks owing to their competitive performance on explicitly modeling label dependencies. However, most of these methods ignore the associations between semantic labels and image local features, and the spatial contexts of images are not sufficiently exploited. Some other works \cite{zhu2017learning} \cite{chen2019learning} introduce attention mechanisms to adaptively search semantic-aware instance regions and aggregate features from these regions to identify multiple labels. However, due to the lack of fine-grained supervision information, these methods could merely locate instance regions roughly, which neither consider the interactions among instances nor explicitly describe the instance-label assignment relationship.

To address the above mentioned issues, in this paper, we fully explore the correspondence (matching) between each instance and label, and reformulate the task of MLIC as an instance-label matching selection problem. Accordingly, we propose a novel Graph Matching based Multi-Label Image Classification (GM-MLIC) deep learning model, which simultaneously incorporates instance spatial relationship, label semantic correlation and co-occurrence possibility of varying instance-label assignments into a unified framework.
Specifically, inspired by Graph Matching scheme (GM) suitable for structured data, we first design an instance spatial graph by representing each instance feature as the node attribute and the relative location relationship of adjacent instances as the edge attribute. Meanwhile, a label semantic graph is employed to capture the overall semantic correlations, which takes the word embedding of each label as node attribute, and concatenates the attributes of two nodes (labels) associated with the same edge to form the edge attribute. Then each instance is connected to all labels to form the final assignment graph, as shown in Figure \ref{ASGC}, which aims to explicitly model the instance-label matching possibility. Furthermore, the Graph Network Block (GNB) is introduced to our framework to perform computation on the constructed assignment graph, which forms structured representations for each instance and label by a graph propagation mechanism. Note that the label co-occurrence relationships in our network are based on learning from the raw attribute of labels and instances, instead of directly adopting the statistics information. Finally, we design a weighted cross-entropy loss to optimize our network output, which indicates the prediction score of each instance-label correspondence. Extensive experiments demonstrate that our proposed method can achieve superior performance against state-of-the-art methods.



\section{Related Work}
The task of MLIC has attracted an increasing interest recently. A straightforward way to address this problem is to train independent binary classifiers for each label. However, such method does not consider the relationship among labels, and the number of predicted labels will grow exponentially as the number of categories increase. To overcome the challenge of such an enormous output space, some works convert multi-label problem into a set of multi-class problems over region proposals. For example, \cite{wei2015hcp} extracted an arbitrary number of object proposals, then aggregated the label confidences of these proposals with max-pooling to obtain the final multi-label predictions.
\cite{yang2016exploit} treated each image as a bag of instances/proposals, and solved the MLIC task in a multi-instance learning manner. 
However, the above methods ignore the label correlation in multi-label images when converting MLIC to the multi-class task. 

Recently, researchers focus on exploiting the label correlation to facilitate the learning process. \cite{gong2013deep} leveraged a ranking-based learning strategy to train deep convolutional neural networks for MLIC and found that the weighted approximated-ranking loss can implicitly model label correlation and work best. \cite{wang2016cnn} utilized recurrent neural networks (RNNs) to transform labels into embedded label vectors, so that the correlation between labels can be employed. Besides, some studies utilize graph structure to model the label correlation. \cite{chen2019multi} used Graph Convolutional Network to map a group of label semantic embeddings borrowed from natural language processing into inter-dependent classifiers.
\cite{wang2020multi} proposed to model label correlation by superimposing label graph built from statistical co-occurrence information into the graph constructed from knowledge priors of labels. However, none of the aforementioned methods consider the associations between semantic labels and image local features, and the spatial contexts of images have not been sufficiently exploited.


To solve the above issues, recent progress on MLIC attempt to model label correlation with region-based multi-label approaches.
For example, \cite{wang2017multi} introduced a spatial transformer to locate semantic-aware instance regions and then captured the spatial dependencies of these regions by Long Short-Term Memory (LSTM).
\cite{chen2019learning} incorporated category semantics to better learn instance features and explored their interactions under the guidance of statistical label dependencies. Although the above methods have achieved competitive performance, they neither consider the spatial location relationships among instances nor explicitly describe the instance-label assignment relationships, which
may make these methods lose the ability to effectively represent the categories visual features. Different from all these methods, in this paper, we utilize the GM scheme and propose a novel multi-label image classification learning framework called GM-MLIC, where the instance spatial relationship, label semantic correlation and instance-label assignment possibility are simultaneously incorporated into the framework to improve the classification performance. The details of the framework are introduced in the following section.


\begin{figure*}[!htbp]
\vspace{-5mm}
\centering  
\centerline{\includegraphics[scale=0.35]{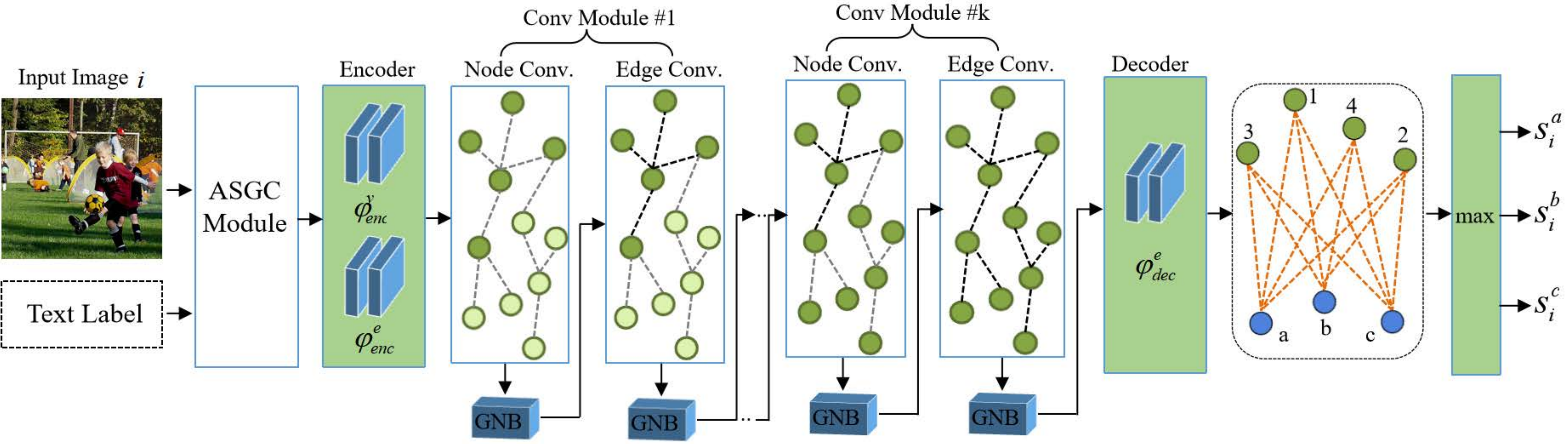}}
\vspace{-3mm}
\caption{Illustration of our proposed deep learning framework for MLIC task. Overall our model consists of two major components: the ASsignment Graph Construction (ASGC) and the Instance-Label Matching Selection (ILMS). The details of ASGC are shown in Figure \ref{ASGC}. The ILMS module consists of Encoder, Graph Convolution and Decoder. The encoder and decoder are designed as MLPs, where encoder transforms all node attributes and edge attributes into latent space and decoder derives instance-label matching scores from the updated graph state. Moreover, the graph convolution utilizes Graph Network Block (GNB) to perform nodes and edges attribute aggregating and updating.}
\label{overview}
\vspace{-5mm}
\end{figure*}

\section{The Proposed Method}
Given a multi-label image dataset $\mathcal{D}=\{(X_i,Y_i)\}_{i=1}^N$, we denote ${X_i}=\{x_i^1,x_i^2,...,x_i^M\}$ as the $i$-th image that consists of $M$ instances, $Y_i=[y_i^1,y_i^2,...,y_i^C]^T$ as the ground-truth label vector for $X_i$, where each instance $x_i^j$ is a $d$-dimensional feature vector and $C$ is the number of all possible label in the dataset. ${y_i^c}=1$ indicates that image $X_i$ is annotated with label $c$, and ${y_i^c}=0$ otherwise. GM-MLIC aims to learn a multi-label classification model from the instance-level feature vector together with image-level ground-truth label vector, and further assign the predictive labels for a test image.

\subsection{Overview}
The overview architecture of the proposed GM-MLIC is illustrated in Figure \ref{overview}, which consists of two components: the ASsignment Graph Construction (ASGC) and the Instance-Label Matching Selection (ILMS). The constructed assignment graph takes instance spatial graph and label semantic graph as input, and considers each instance-label connection as candidate matching edge. The matching selection module is another core component of our learning framework, which introduces Graph Network Block (GNB) to convolve the information of neighborhoods of each instance and label to form a structured representation through several convolution operators. Finally, our model derives a prediction score for each instance-label correspondence and optimizes such correspondence with a weighted cross-entropy loss.

\subsection{The Assignment Graph Construction}
As depicted in Figure \ref{ASGC}, we first construct an instance spatial graph to explore the relationship of the spatially adjacent instances. Specifically, we feed an image to a pre-trained Faster R-CNN network \cite{ren2016faster} to generate a set of semantic-aware instances, where each instance contains a bounding box $B(x,y,w,h)$. Within the bounding box, $(x,y)$ denotes the location ($x$-axis, $y$-axis coordinates) and $(w,h)$ denotes the size (width and height). Each of these instances is taken as a node in the instance spatial graph $\mathbb{G}^o$ in which the edges between each pair of nodes are produced through $k$-Nearest Neighbor criteria. Note that our $\mathbb{G}^o$ is a directed graph, where the attributes of nodes are denoted by the feature $f$ of corresponding instances, and the attributes of edges are represented by the concatenations of bounding box coordinates of its source instance and receive instance
\begin{equation}
v_i^{o}=f_i,
\hspace{1em}
e_{ij}^{o}=[\hat{B}_i,\hat{B}_j],
\end{equation}
where $f_i$ and $\hat{B}_{i}(x_i,y_i,x_i+w_i,y_i+h_i)$ denote the attributes and location coordinates of $i$-th instance respectively and $[\cdot,\cdot]$ the concatenations of its input.

Similar to \cite{chen2019multi} \cite{you2020cross}, we construct a label semantic graph $\mathbb{G}^l$ to capture the topological structure in the label space, where each node of the graph is represented as word embeddings of the label. Different from above these methods, our $\mathbb{G}^l$ does not require pre-defined label co-occurrence statistics. It instead combines the word embeddings of the connected two nodes (labels) to form the initial edges attributes of $\mathbb{G}^l$ as
\begin{equation}
v_i^{l}=w_i,
\hspace{1em}
e_{ij}^{l}=[w_i,w_j],
\end{equation}
where $w_i$ denotes the word embedding of $i$-th label.

To explicitly establish the instance-label matching relationship, we connect each instance in $\mathbb{G}^o$ to all labels in $\mathbb{G}^l$ to form the instance-label assignment graph $\mathbb{G}^A$. In $\mathbb{G}^A$, the attributes of the matching edge that connecting instance and label are represented as
\begin{equation}
e_{ij}^{m}=[v_i^{o},v_j^{l}].
\end{equation}

In this way, we successfully convert the problem of building the complex correspondence between an image and its labels to the issue of selecting reliable edges from a constructed assignment graph. Accordingly, the goal of the MLIC problem is transformed into how to solve the matching selection problem and obtain the optimal instance-label assignment.

\subsection{Modeling Instance-Label Correspondence}
As illustrated in Figure \ref{overview}, our matching selection module is designed on the top of Graph Network Block (GNB) presented in \cite{wang2020learning}, which defines a class of functions for relation reasoning over graph-structured representations. It is extended in our model to fit the instance-label matching selection problem by eliminating redundant components and redefining certain functions. Specifically, the matching selection module consists of three main components: Encoder, Graph Convolution and Decoder.

\textbf{Encoder:} The encoder takes the constructed assignment graph $\mathbb{G}^A$ as input, and transforms its attributes into a latent feature space by two parametric update functions $\varphi_{enc}^{v}$ and $\varphi_{enc}^{e}$. In our framework, $\varphi_{enc}^{v}$ and $\varphi_{enc}^{e}$ are designed as multi-layer perceptrons (MLPs), each of which takes respectively a node attribute vector and an edge attribute vector as input and transforms them into latent spaces. Note that $\varphi_{enc}^{v}$ is mapped across all nodes (instances and label) to compute per-node updates, and $\varphi_{enc}^{e}$ is mapped across all edges (relationships) to compute per-edge updates.

Formally speaking, we denote $\varphi_{enc}^{v}(\mathcal{V}^{A})$ and $\varphi_{enc}^{e}(\mathcal{E}^{A})$ as the updated node attributes and edge attributes by applying $\varphi_{enc}^{v}$ and $\varphi_{enc}^{e}$ to each node and each edge respectively. Then the encoder module can be briefly described as
\begin{equation}
\mathbb{G}^A \leftarrow Enc(\mathbb{G}^A)=(\mathbb{V}^A,\mathbb{E}^A,\varphi_{enc}^{v}(\mathcal{V}^{A}),\varphi_{enc}^{e}(\mathcal{E}^{A})).
\end{equation}

The updated graph $\mathbb{G}^A$ is then passed to the subsequent convolution modules as input.

\textbf{Graph Convolution Module:} This module consists of a node convolution layer and an edge convolution layer. The node convolution layer collects the attributes of all the nodes and edges adjacent to each node to compute per-node updates. It is followed by the edge convolution layer that assembles the attributes of the two nodes associated with each edge to generate a new attribute of this edge.

Specifically, for the $i$-th node in $\mathbb{G}^o$, the aggregation function gathers the information from its adjacent nodes and associated edges, and the update function outputs the updated attributes according to the gathered information. Given that for each instance node $v_i^{o}$, other instance nodes and label nodes are connected with instance edges and matching edges respectively, we design two types of aggregation functions as
\begin{equation}
\hat{v}_i^{o}=\frac{1}{\|\mathbb{N}^{o}\|}\hat{\rho}_{n}^{o}([e_{ij}^{o},v_j^{o}]),
\hspace{1em}
\tilde{v}_i^{o}=\frac{1}{\|\mathbb{N}^{l}\|}\tilde{\rho}_{n}^{o}([e_{ij}^{c},v_j^{l}]),
\end{equation}
where $\mathbb{N}^{o}$ and $\mathbb{N}^{l}$ are two sets of instance nodes and label nodes adjacent with $v_i^{o}$. $\hat{\rho}_{n}^{o}$ and $\tilde{\rho}_{n}^{o}$ gather the information from object nodes and label nodes respectively.
Then an update function is employed to update the attributes for $v_i^{o}$
\begin{equation}
v_i^{o}\leftarrow \phi^{o}_{n}([v_i^{o},\hat{v}_i^{o},\tilde{v}_i^{o}]),
\end{equation}
where $\phi^{o}_{n}$ takes the concatenation of the current attributes of $v_i^{o}$ and gathered information $\hat{v}_i^{o}$ and $\tilde{v}_i^{o}$, and outputs the updated attributes for $v_i^{o}$.

Similar to instance nodes, the label nodes are connected with two types of nodes and associated with two types of edges. Therefore, we also design two aggregation functions and an update function for label node $v_i^{l}$. Specifically, the aggregation functions are formulated as
\begin{equation}
\hat{v}_i^{l}=\frac{1}{\|\mathbb{N}^{l}\|}\hat{\rho}_{n}^{l}([e_{ij}^{l},v_j^{l}]),
\hspace{1em}
\tilde{v}_i^{l}=\frac{1}{\|\mathbb{N}^{o}\|}\tilde{\rho}_{n}^{o}([e_{ij}^{c},v_j^{o}]),
\end{equation}
and the update function is represented as
\begin{equation}
v_i^{l}\leftarrow \phi^{l}_{n}([v_i^{l},\hat{v}_i^{l},\tilde{v}_i^{l}]).
\end{equation}

For an instance edge $e_{ij}^{o}$, both of its source node and receive node are in $\mathbb{G}^o$. Therefore, we design an aggregation function and the update function as follows
\begin{equation}
\hat{e}_{ij}^{o}=\rho_{e}^{o}([v_i^{o},v_j^{o}]),
\hspace{1em}
{e}_{ij}^{o}\leftarrow \phi^{o}_{e}([{e}_{ij}^{o},\hat{e}_{ij}^{o}]),
\end{equation}
where $\rho_{e}^{o}$ aggregates the information from source instance node $v_i^{o}$ and receive node $v_j^{o}$, $\phi^{o}_{e}$ updates the attributes of ${e}_{ij}^{o}$ according to the gathered information.

Similar to the instance edge convolution operator, the label edge convolution layer consists of an aggregation function and an update function, which are designed as
\begin{equation}
\hat{e}_{ij}^{l}=\rho_{e}^{l}([v_i^{l},v_j^{l}]),
\hspace{1em}
{e}_{ij}^{l}\leftarrow \phi^{l}_{e}([{e}_{ij}^{l},\hat{e}_{ij}^{l}]).
\end{equation}

Different from instance edges and label edges, the matching edges connect instance nodes and label nodes. Therefore, the aggregation function gathers information from different type of nodes
\begin{equation}
\hat{e}_{ij}^{m}=\rho_{e}^{m}([v_i^{o},v_j^{l}]),
\end{equation}
and the update function also takes the combination of aggregated features and its current features as input and output the updated attributes
\begin{equation}
{e}_{ij}^{m}\leftarrow \phi^{m}_{e}([{e}_{ij}^{m},\hat{e}_{ij}^{m}]).
\end{equation}

All the above aggregation functions and update functions are designed as MLPs, but their structure and parameters are different from each other.

 \textbf{Decoder:} The decoder module reads out the final output from the updated graph state. Since only the attributes of instance-label matching edges are required for final evaluation, the decoder module contains only one update function $\varphi_{dec}^{e}$ that transforms the edges attributes into the desired space
\begin{equation}
S=Dec(\mathbb{G}^A)=\varphi_{dec}^{e}(\mathcal{E}^{A}),
\end{equation}
where $S\in {[0,1]}^{M\times C}$ denotes the prediction score that each instance is matched with the corresponding label. Similarly, $\varphi_{enc}^{e}$ is parameterized by an MLP.

\subsection{Optimizing Multi-Label Prediction}
In order to interpret each ground truth label of the input image, there should be at least one instance that best matches it. With consideration of the possibly noisy instances, a cross-instances max-pooling is carried out to fuse the output of our framework into an integrative prediction. Suppose $s_j(j=1,2,...,M)$ is the prediction score vector of the $j$-th instance from the decoder and $s_j^c(c=1,2,...,C)$ is the $c$-th category matching score of $s_j$. The cross-instances max-pooling can be formulated as
\begin{equation}
p_i^c=max(s_1^c,s_2^c,...,s_M^c),
\end{equation}
where $p_i^c$ can be considered as the prediction score for the $c$-th category of the given image $i$.

Finally, the training process of our network is guided by a weighted cross-entropy loss with the ground-truth labels $Y_i$ as supervision,
\begin{gather}
 \mathcal{L}=\sum_{i=1}^{N}\sum_{c=1}^{C}w^c[y_i^c\log(p_i^c)+(1-y_i^c)\log(1-p_i^c))]\nonumber\\
 w^c=y_i^c\cdot e^{\beta(1-r^c)}+(1-y_i^c)\cdot e^{\beta r^c}, \label{loss}
 \end{gather}
where $w^c$ is used to alleviate the class imbalance, $\beta$ is a hyper-parameter and $r^c$ is the ratio of label $c$ in the training set.

\begin{table*}[htbp]
\vspace{-5mm}
\begin{center}
\small
\setlength{\tabcolsep}{0.2mm}{
\begin{tabular}{c|c|c|c|c|c|c|c|c|c|c|c|c|c|c|c|c|c|c|c|c|c}
\toprule[1pt]
Methods &\emph{aero} &\emph{bike} &\emph{bird} &\emph{boat} &\emph{bottle} &\emph{bus} &\emph{car} &\emph{cat} &\emph{chair} &\emph{cow} &\emph{table} &\emph{dog} &\emph{horse} &\emph{motor} &\emph{person} &\emph{plant} &\emph{sheep} &\emph{sofa} &\emph{train} &\emph{tv} &mAP \\\hline
HCP \cite{wei2015hcp} &98.6 &97.1 &\textcolor{blue}{98.0} &95.6 &75.3 &94.7 &95.8 &97.3 &73.1 &90.2 &80.0 &97.3 &96.1 &94.9 &96.3 &78.3 &94.7 &76.2 &97.9 &91.5 &90.9 \\
CNN-RNN \cite{wang2016cnn} &96.7 &83.1 &94.2 &92.8 &61.2 &82.1 &89.1 &94.2 &64.2 &83.6 &70.0 &92.4 &91.7 &84.2 &93.7 &59.8 &93.2 &75.3 &99.7 &78.6 &84.0 \\
ResNet-101 \cite{he2016deep} &\textcolor{blue}{99.5} &97.7 &97.8 &96.4 &65.7 &91.8 &96.1 &97.6 &74.2 &80.9 &85.0 &\textcolor{blue}{98.4} &96.5 &95.9 &98.4 &70.1 &88.3 &80.2 &\textcolor{blue}{98.9} &89.2 &89.9 \\
RNN-Attention \cite{wang2017multi} &98.6 &97.4 &96.3 &96.2 &75.2 &92.4 &96.5 &97.1 &76.5 &92.0 &\textcolor{blue}{87.7} &96.8 &97.5 &93.8 &98.5 &81.6 &93.7 &82.8 &98.6 &89.3 &91.9 \\
ML-GCN \cite{chen2019multi} &\textcolor{red}{99.6} &98.3 &97.9 &\textcolor{blue}{97.6} &78.2 &92.3 &\textcolor{blue}{97.4} &97.4 &79.2 &94.4 &86.5 &97.4 &97.9 &\textcolor{blue}{97.1} &98.7 &84.6 &95.3 &83.0 &98.6 &90.4 &93.1 \\
SSGRL \cite{chen2019learning} &\textcolor{blue}{99.5} &97.1 &97.6 &\textcolor{blue}{97.8} &82.6 &\textcolor{blue}{94.8} &96.7 &\textcolor{blue}{98.1} &78.0 &\textcolor{blue}{97.0} &85.6 &97.8 &\textcolor{blue}{98.3} &96.4 &\textcolor{blue}{98.8} &\textcolor{blue}{84.9} &96.5 &79.8 &98.4 &92.8 &93.4 \\
TSGCN \cite{xu2020joint} &98.9 &\textcolor{blue}{98.5} &96.8 &97.3 &\textcolor{red}{87.5} &94.2 &\textcolor{blue}{97.4} &97.7 &\textcolor{red}{84.1} &92.6 &\textcolor{red}{89.3} &\textcolor{blue}{98.4} &98.0 &96.1 &98.7 &\textcolor{blue}{84.9} &\textcolor{blue}{96.6} &\textcolor{red}{87.2} &98.4 &\textcolor{blue}{93.7} &\textcolor{blue}{94.3} \\\hline
GM-MLIC &99.4 &\textcolor{red}{98.7} &\textcolor{red}{98.5} &\textcolor{blue}{97.6} &\textcolor{blue}{86.3} &\textcolor{red}{97.1} &\textcolor{red}{98.0} &\textcolor{red}{99.4} &\textcolor{blue}{82.5} &\textcolor{red}{98.1} &\textcolor{blue}{87.7} &\textcolor{red}{99.2} &\textcolor{red}{98.9} &\textcolor{red}{97.5} &\textcolor{red}{99.3} &\textcolor{red}{87.0} &\textcolor{red}{98.3} &\textcolor{blue}{86.5} &\textcolor{red}{99.1} &\textcolor{red}{94.9} &\textcolor{red}{94.7}\\
\bottomrule[1pt]
\end{tabular}}
\vspace{-3mm}
\caption{Comparisons of AP and mAP with state-of-the-art methods on the VOC 2007 dataset.The \textcolor{red}{red} numbers indicate the best results in different metrics, while the \textcolor{blue}{blue} numbers indicate the sub-optimal results. Best viewed in color.}
\label{voc2007}
\end{center}
\end{table*}

\begin{table*}[htbp]
\vspace{-3mm}
\begin{center}
\setlength{\tabcolsep}{1.9mm}{
\begin{tabular}{c|c|c|c|c|c|c|c|c|c|c|c|c|c}
\toprule[1pt]
\multirow{2}{*}{Methods} &\multicolumn{7}{c|}{All} &\multicolumn{6}{c}{Top-3}  \\\cline{2-14}
                         & mAP & CP & CR & CF1 & OP & OR & OF1 & CP & CR & CF1 & OP & OR & OF1  \\\hline\hline
CNN-RNN \cite{wang2016cnn} &61.2 &- &- &- &- &- &- &66.0 &55.6 &60.4 &69.2 &66.4 &67.8 \\
ResNet-101 \cite{he2016deep} &77.3 &80.2 &66.7 &72.8 &83.9 &70.8 &76.8 &84.1 &59.4 &69.7 &89.1 &62.8 &73.6 \\
RNN-Attention \cite{wang2017multi} &- &- &- &- &- &- &- &79.1 &58.7 &67.4 &84.0 &63.0 &72.0 \\
SRN \cite{zhu2017learning} &77.1 &81.6 &65.4 &71.2 &82.7 &69.9 &75.8 &85.2 &58.8 &67.4 &87.4 &62.5 &72.9 \\
ML-GCN \cite{chen2019multi} &83.0 &85.1 &72.0 &\textcolor{blue}{78.0} &85.8 &\textcolor{blue}{75.4} &\textcolor{blue}{80.3} &89.2 &64.1 &\textcolor{blue}{74.6} &90.5 &66.5 &76.7 \\
SSGRL \cite{chen2019learning} &\textcolor{blue}{83.6} &\textcolor{red}{89.5} &68.3 &76.9 &\textcolor{red}{91.2} &70.7 &79.3 &\textcolor{red}{91.9} &62.1 &73.0 &\textcolor{blue}{93.6} &64.2 &76.0 \\
CMA \cite{you2020cross} &83.4 &82.1 &\textcolor{red}{73.1} &77.3 &83.7 &\textcolor{red}{76.3} &79.9 &87.2 &64.6 &74.2 &89.1 &66.7 &76.3 \\
TSGCN \cite{xu2020joint} &83.5 &81.5 &\textcolor{blue}{72.3} &76.7 &84.9 &75.3 &79.8 &84.1 &\textcolor{blue}{67.1} &\textcolor{blue}{74.6} &89.5 &\textcolor{blue}{69.3} &\textcolor{red}{78.1} \\\hline
GM-MLIC &\textcolor{red}{84.3} &\textcolor{blue}{87.3} &70.8 &\textcolor{red}{78.3} &\textcolor{blue}{88.6} &74.8 &\textcolor{red}{80.6} &\textcolor{blue}{90.6} &\textcolor{red}{67.3} &\textcolor{red}{74.9} &\textcolor{red}{94.0} &\textcolor{red}{69.8} &\textcolor{blue}{77.8}\\
\bottomrule[1pt]
\vspace{-6mm}
\end{tabular}}
\caption{ Comparisons with state-of-the-art methods on the MS-COCO dataset.The \textcolor{red}{red} numbers indicate the best results in different metrics, while the \textcolor{blue}{blue} numbers indicate the sub-optimal results. Best viewed in color. '-' indicates the results are not reported in the original literature.}
\label{coco}
\end{center}
\vspace{-5mm}
\end{table*}

\section{Experiments}
\subsection{Evaluation Metrics}
We adopt six widely used multi-label metrics to evaluate each comparing method, including the average per-class precision (CP), recall (CR), F1 (CF1) and the average overall precision (OP), recall (OR), F1 (OF1), whose detailed definitions can be found in \cite{chen2019learning}. In this paper, we present the above metrics under the setting that a label is predicted as positive if its estimated probability is greater than 0.5. To fairly compare with the state-of-the-art methods, we also report the results of top-3 labels. Besides, we compute and report the average precision (AP) and mean average precision (mAP).

\subsection{Implementation Details}
In ASGC module, we apply Faster R-CNN (resnet50-fpn) \cite{ren2016faster} to generate a set of instances for per image, where each of these instances, in addition to the visual feature $f$ and bounding box $B$, also contains a preliminary class label $c$ with a confidence score $s$. To save computational cost, we only select the instances with the top-$m$ confidence score for each image. For label representations, we adopt 300-dim GloVe \cite{pennington2014glove} trained on the Wikipedia dataset. In ILMS module, the Graph Convolution Layer consists of $k$ convolution modules, where they are stacked to aggregate the information of $k^{th}$-order neighborhoods. In our experiments, $k$ is 2 and the output dimension of the corresponding convolution module is 512 and 256, respectively. During training, the input images are randomly cropped and resized into 448$\times$ 448 with random horizontal flips for data augmentation. All modules are implemented in PyTorch and the optimizer is SGD with momentum 0.9. Weight decay is 10$^{-4}$. The initial learning rate is 0.01, which decays by a factor of 10 for every 30 epochs. And the hyperparameter $\beta$ in the Eq. (\ref{loss}) is set to 0 in VOC 2007 dataset and 0.4 in both MS-COCO and NUS-WIDE datasets.

\subsection{Comparisons with State-of-the-Arts}

\textbf{Results on VOC 2007:}
Pascal VOC 2007 \cite{everingham2010pascal} is the most widely used dataset to evaluate the MLIC task, which covers 20 common categories and contains a trainval set of 5,011 images and a test set of 4,952 images. We present the AP of each category and mAP over all categories on the VOC 2007 dataset in Table \ref{voc2007}. Considering that the dataset is less complicated and its size is relatively small, our proposed model still achieves 94.7\% mAP, which is respectively 0.4\%, 1.3\% and 1.6\% superior over TSGCN, SSGRL and ML-GCN on VOC 2007 dataset. Particularly, our GM-MLIC obtains significant improvements in the categories of small objects, including \textit{bird}, \textit{plant}, \textit{sheep} and \textit{cat} with 98.5\%, 87.0\%, 98.3\% and 99.4\% in terms of AP.

\begin{figure*}[!ht]
\vspace{-5mm}
\centering
\begin{tabular}{cccccc}
\vspace{-1mm}
\includegraphics[width=28mm,height=18mm]{figure03.pdf}\hspace{-6mm}
&\includegraphics[width=28mm,height=18mm]{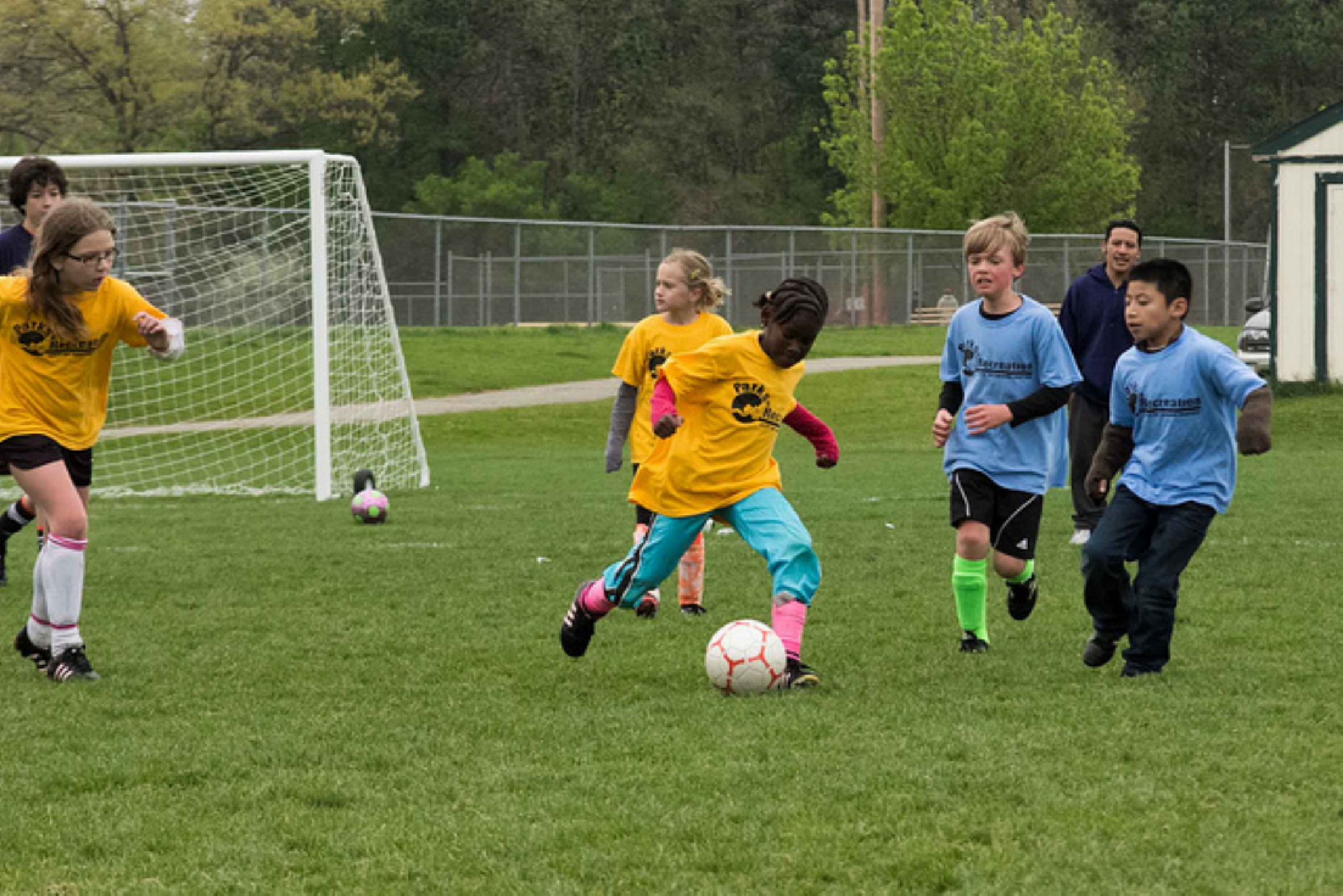}\hspace{-6mm}
&\includegraphics[width=28mm,height=18mm]{figure05.pdf}\hspace{-3mm}
&\includegraphics[width=28mm,height=18mm]{figure06.pdf}\hspace{-3mm}
&\includegraphics[width=28mm,height=18mm]{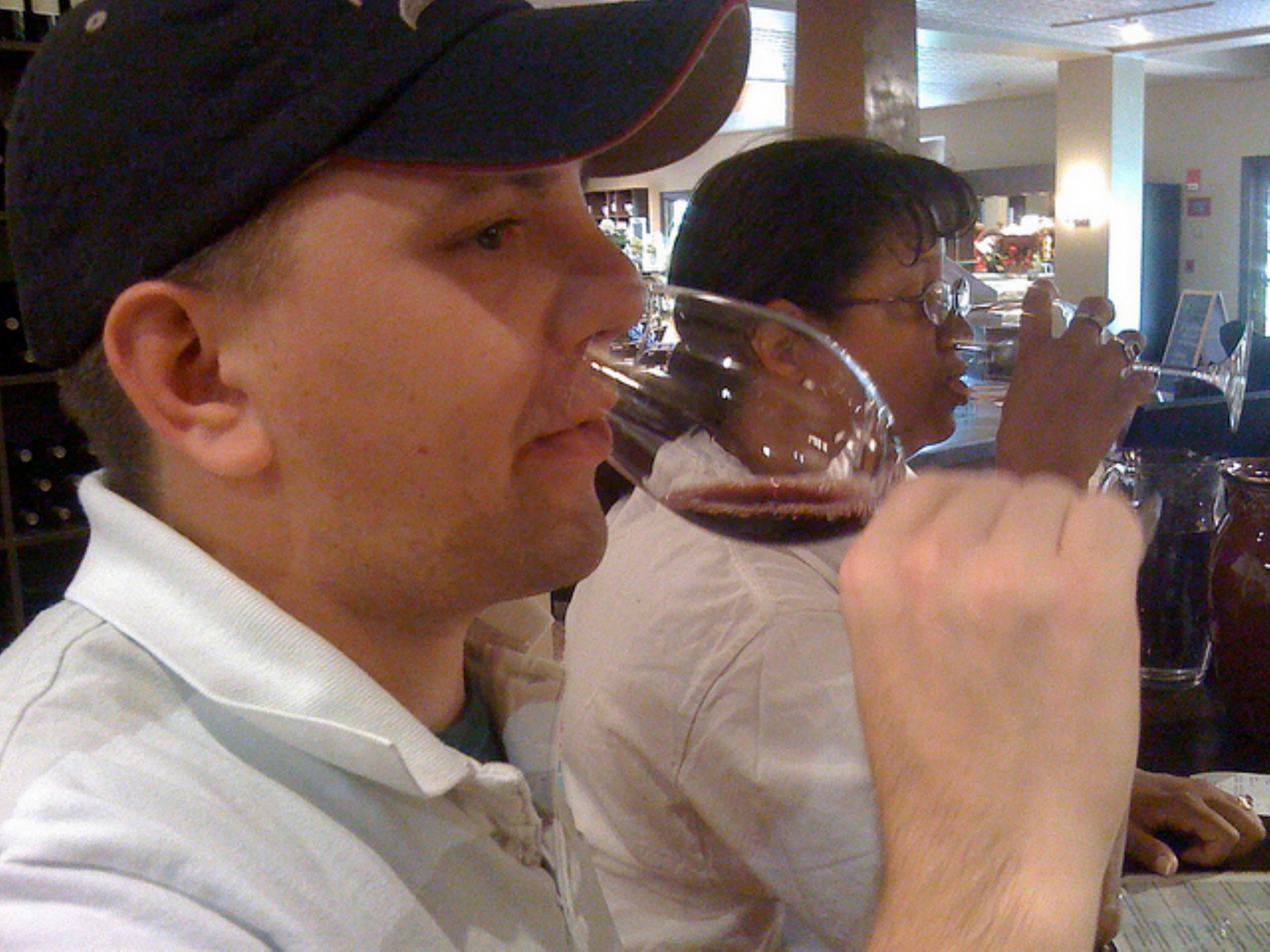}\hspace{-3.5mm}
&\includegraphics[width=28mm,height=18mm]{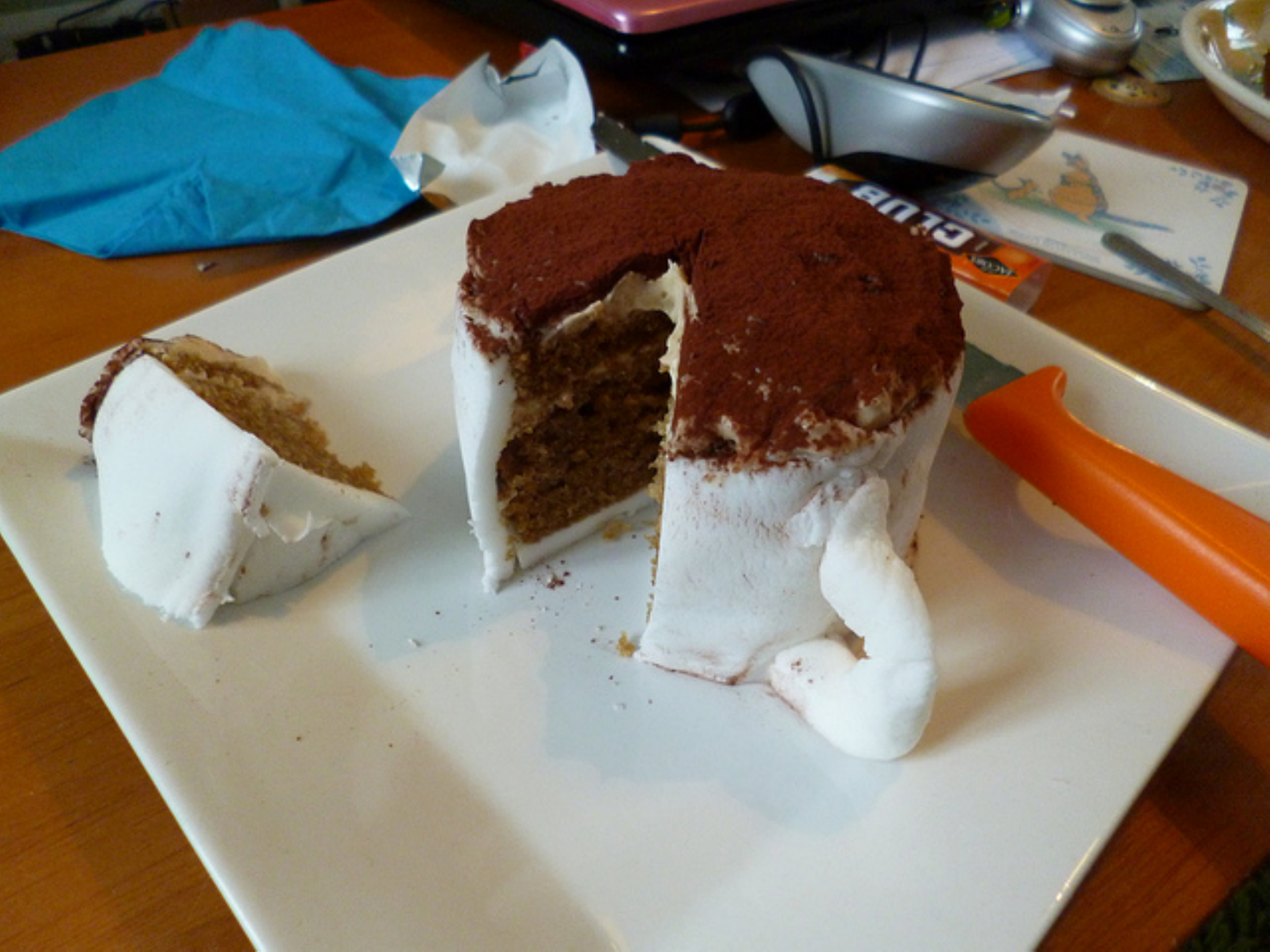}\\
\textcolor{red}{\scriptsize{car,person,bus}}\hspace{-6mm}
&\textcolor{red}{\scriptsize{person,sports ball,bench}}\hspace{-6mm}
&\textcolor{red}{\scriptsize{bowl,banana,person}}\hspace{-3mm}
&\textcolor{red}{\scriptsize{person,frisbee,umbrella}}\hspace{-3mm}
&\textcolor{red}{\scriptsize{person,wine glass,vase}}\hspace{-3.5mm}
&\textcolor{red}{\scriptsize{cake,knife,mouse}} \\
\textcolor{blue}{\scriptsize{car,truck,stop sign}}\hspace{-6mm}
&\textcolor{blue}{\scriptsize{person,sports ball,bottle}}\hspace{-6mm}
&\textcolor{blue}{\scriptsize{cake,person,dining table}}\hspace{-3mm}
&\textcolor{blue}{\scriptsize{person,frisbee,chair}}\hspace{-3mm}
&\textcolor{blue}{\scriptsize{person,donut,laptop}}\hspace{-3.5mm}
&\textcolor{blue}{\scriptsize{cake,dining,table,knife}} \\
\vspace{-1mm}
\includegraphics[width=28mm,height=18mm]{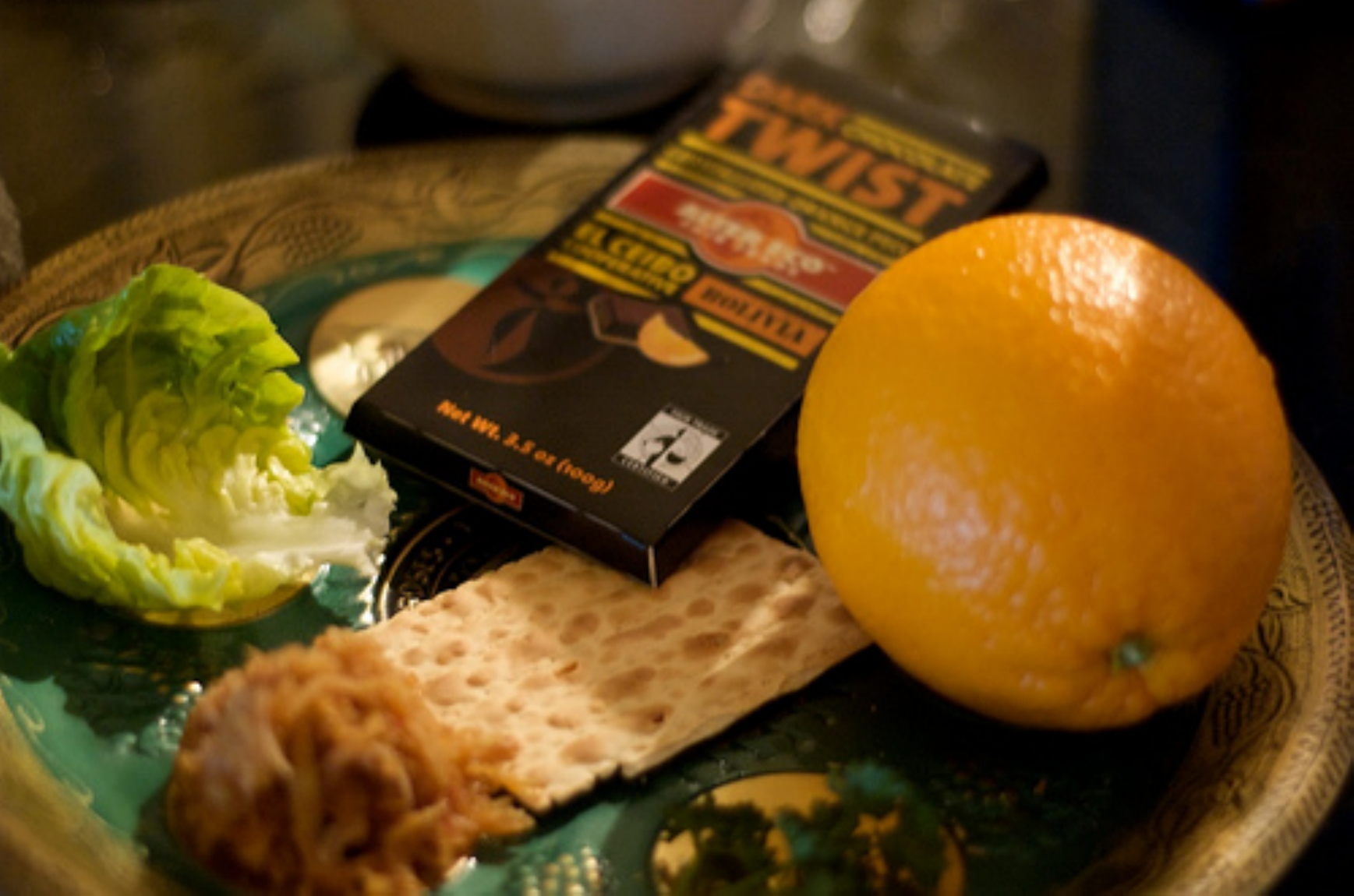}\hspace{-6mm}
&\includegraphics[width=28mm,height=18mm]{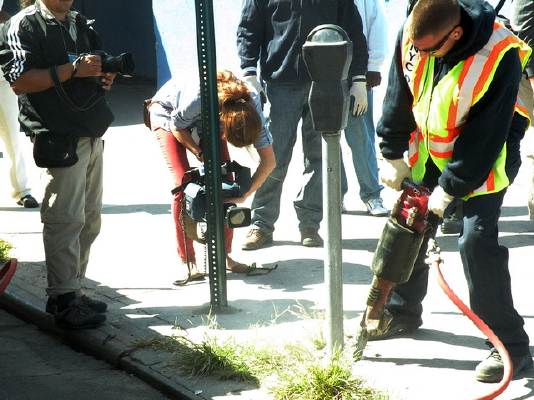}\hspace{-6mm}
&\includegraphics[width=28mm,height=18mm]{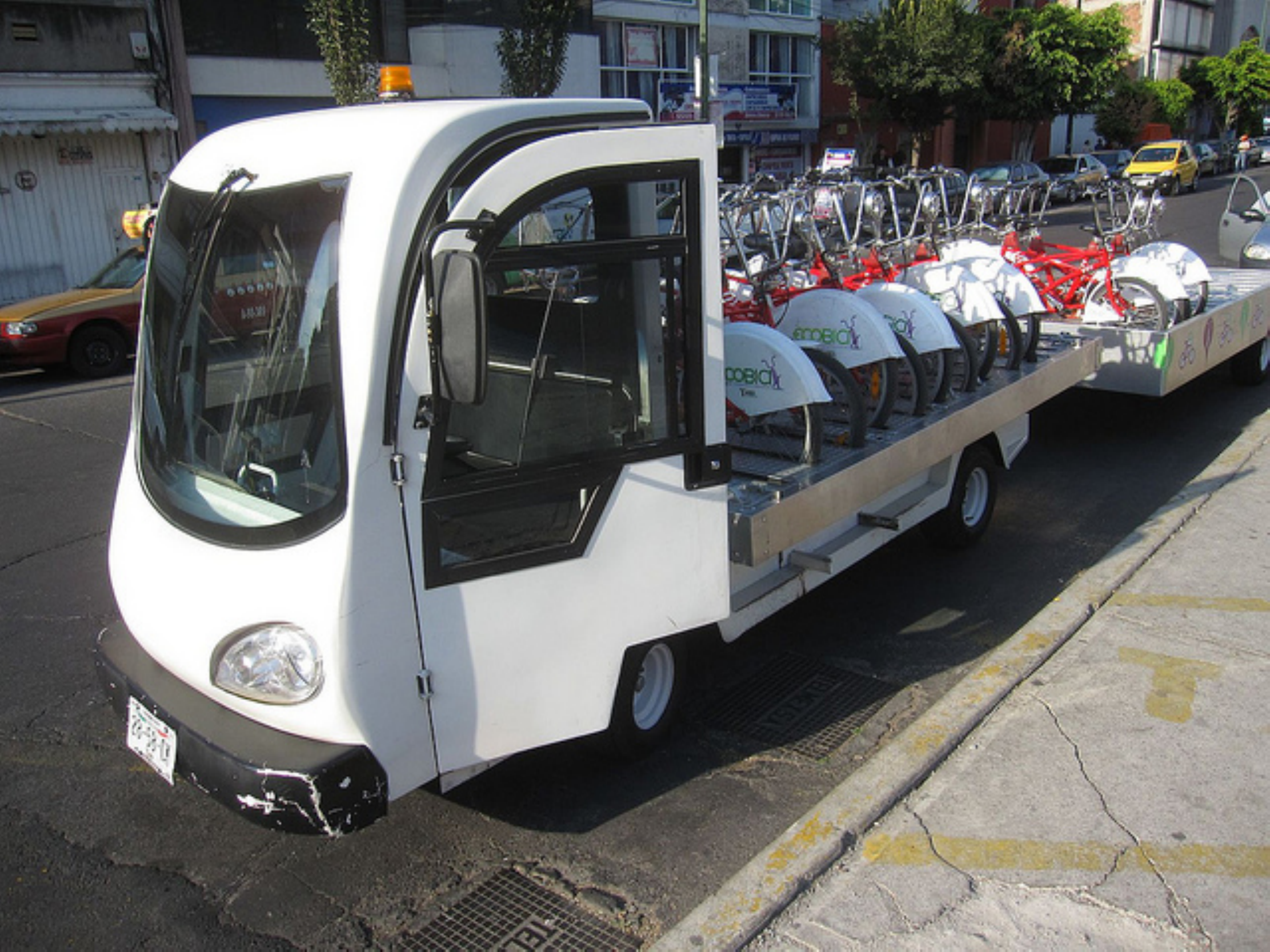}\hspace{-3mm}
&\includegraphics[width=28mm,height=18mm]{figure12.pdf}\hspace{-3mm}
&\includegraphics[width=28mm,height=18mm]{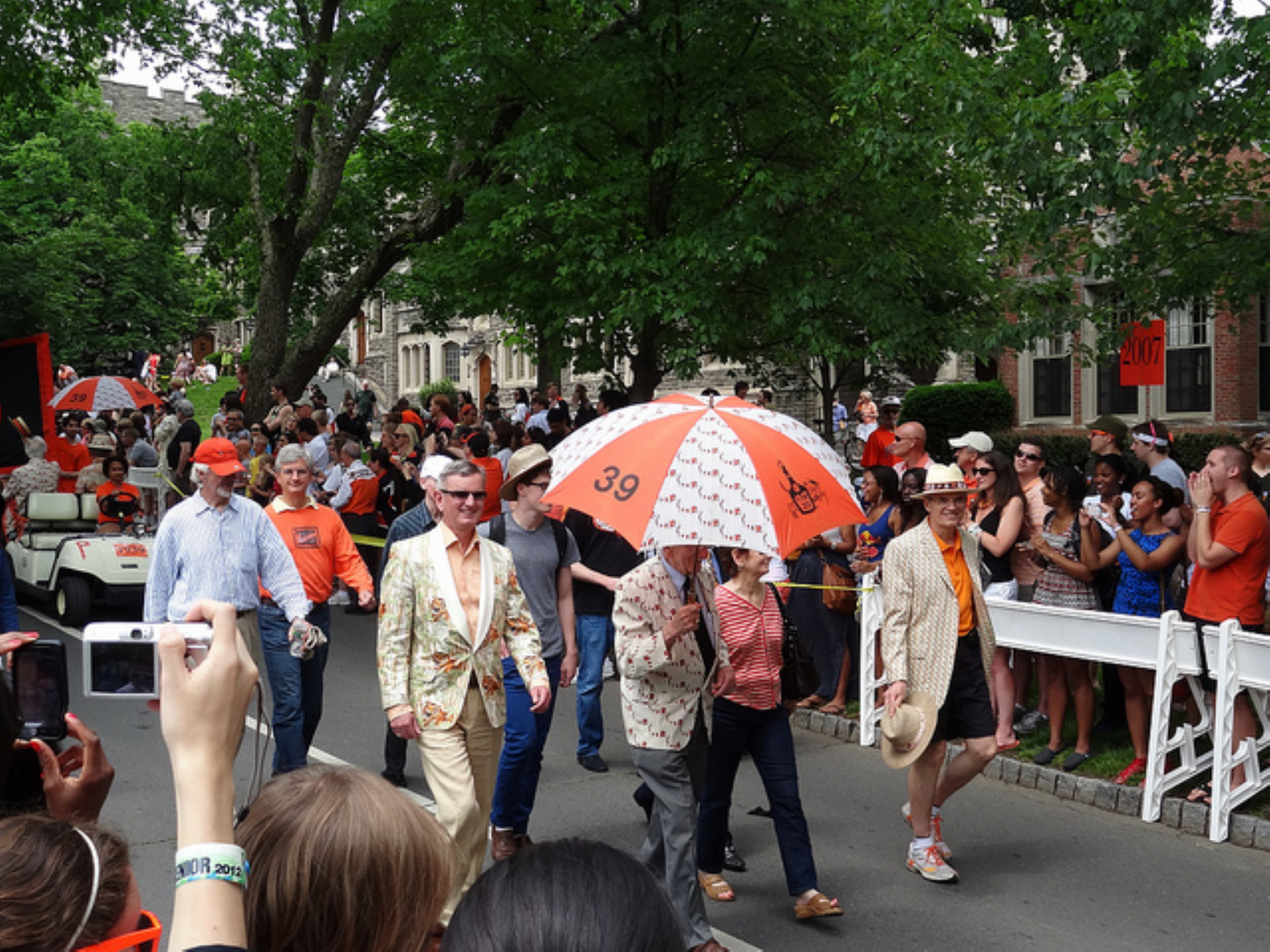}\hspace{-3.5mm}
&\includegraphics[width=28mm,height=18mm]{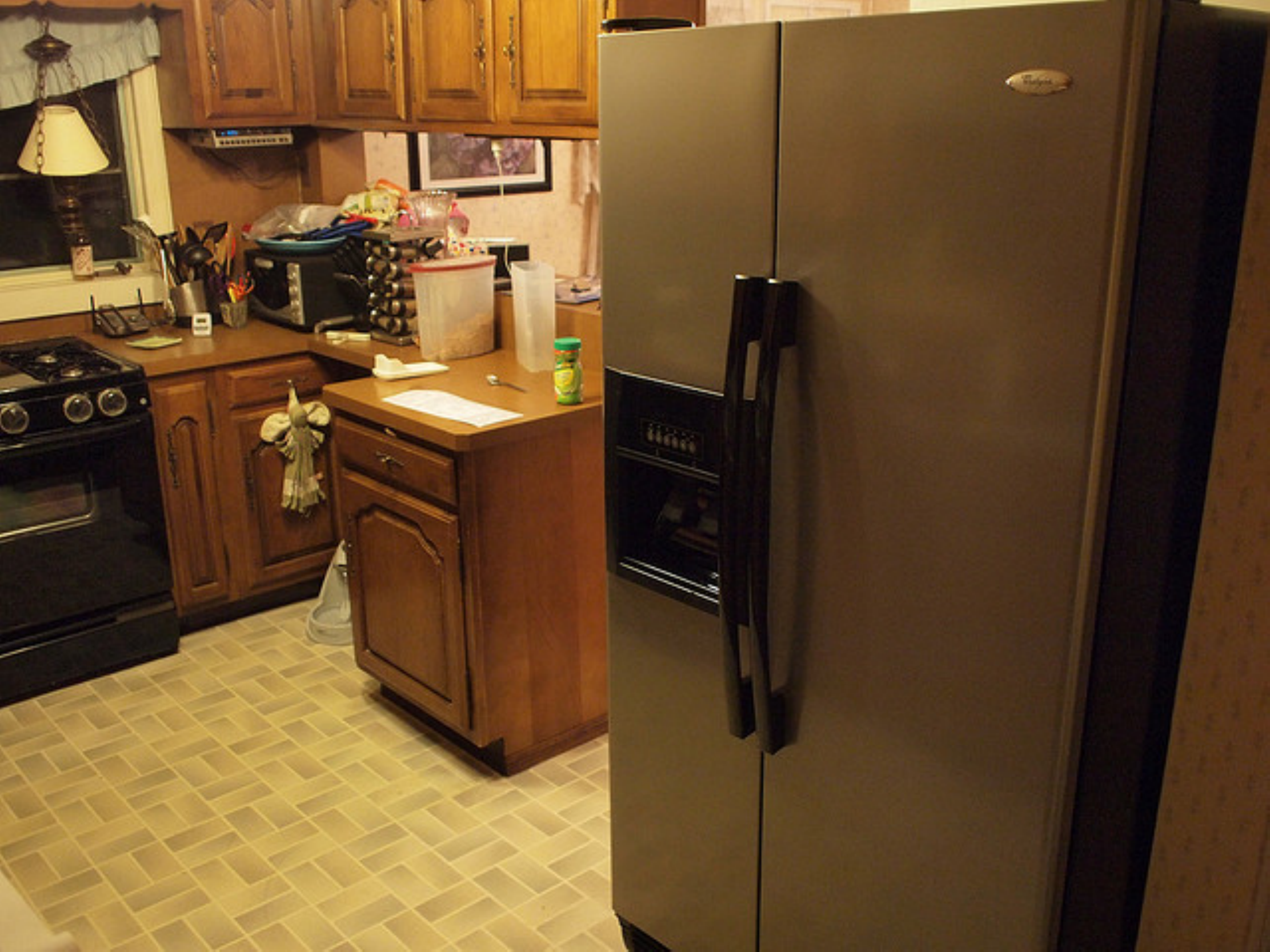}\\
\textcolor{red}{\scriptsize{orange,bowl,dining table}} \hspace{-6mm}
&\textcolor{red}{\scriptsize{person,handbag,parking meter}}\hspace{-6mm}
&\textcolor{red}{\scriptsize{truck,bicycle,car}} \hspace{-3mm}
&\textcolor{red}{\scriptsize{person,frisbee,umbrella}} \hspace{-3mm}
&\textcolor{red}{\scriptsize{pizza,fork,spoon}} \hspace{-3.5mm}
&\textcolor{red}{\scriptsize{spoon,microwave,refrigerator}}\\
\textcolor{blue}{\scriptsize{orange,bowl,book}} \hspace{-6mm}
&\textcolor{blue}{\scriptsize{person,parking meter,potted plant}} \hspace{-6mm}
&\textcolor{blue}{\scriptsize{car,umbrella,truck}} \hspace{-3mm}
&\textcolor{blue}{\scriptsize{person,frisbee,chair}} \hspace{-3mm}
&\textcolor{blue}{\scriptsize{dining table,chair,sandwich}} \hspace{-3.5mm}
&\textcolor{blue}{\scriptsize{refrigerator,oven,cup}}\\
\end{tabular}
\vspace{-3mm}
\caption{Top-3 returned labels by GM-MLIC (\textcolor{red}{red} word) and SSGRL(\textcolor{blue}{blue} word). Best viewed in color.}
\label{visual}
\vspace{-4mm}
\end{figure*}

\begin{table}[t]
\begin{center}
\setlength{\tabcolsep}{0.28mm}{
\begin{tabular}{c|c|c|c|c|c}
\toprule[1pt]
\multirow{2}{*}{Methods} &\multicolumn{3}{c|}{All} &\multicolumn{2}{c}{Top-3} \\ \cline{2-6}
               & mAP & CF1 & OF1 &CF1 &OF1  \\\hline

CNN-RNN \cite{wang2016cnn} &56.1 &- &- &34.7 &55.2  \\
SRN \cite{zhu2017learning} &\textcolor{blue}{61.8} &56.9 &73.2 &47.7 &62.2  \\
MLIC-KD-WSD \cite{liu2018multi} &60.1 &58.7 &\textcolor{blue}{73.7} &53.8 &71.1  \\
PLA \cite{yazici2020orderless} &- &56.2 &- &- &\textcolor{blue}{72.3} \\
CMA \cite{you2020cross} &61.4 &\textcolor{blue}{60.5} &\textcolor{blue}{73.7} &\textcolor{red}{55.5} &70.0 \\\hline
GM-MLIC &\textcolor{red}{62.2} &\textcolor{red}{61.0} &\textcolor{red}{74.1} &\textcolor{blue}{55.3} &\textcolor{red}{72.5} \\
\bottomrule[1pt]
\end{tabular}}
\vspace{-3mm}
\caption{ Comparisons with state-of-the-art methods on the NUS-WIDE dataset. The best and suboptimal results are highlighted in \textcolor{red}{red} and \textcolor{blue}{blue}, respectively. Best viewed in color. '-' indicates the results are not reported in the original literature.}
\label{nus-wide}
\end{center}
\vspace{-7mm}
\end{table}

\textbf{Results on MS-COCO:}
Microsoft COCO \cite{lin2014microsoft} contains a training set of 82,081 images and a validation set of 40,137 images, and covers 80 common categories with 2.9 instance labels per image. The number of labels of different images also varies considerably, which makes MS-COCO more challenging. As shown in Table \ref{coco}, GM-MLIC outperforms all baselines in terms of all evaluation metrics in most cases. Specifically, GM-MLIC is superior to other comparing methods with 84.3\% mAP, 78.3\% (74.9\%) CF1 (Top-3), 80.6\% OF1, 67.3\% Top-3 CR and 69.8\% Top-3 OR, respectively. Noted that our proposed method offers significant improvements on recall score and F1 score. we attribute such improvement to the innovative design of the instance spatial graph and instance-label assignment relationship. To visually understand the effectiveness of our model, we randomly select some images from different scenes and exhibit the top-3 returned labels by GM-MLIC and SSGRL in Figure \ref{visual}.


\textbf{Results on NUS-WIDE:} The NUS-WIDE dataset \cite{chua2009nus} contains 161,789 images for training and 107,859 images for testing. The dataset is manually annotated by 81 concepts, with 2.4 concept labels per image on average. Experimental results on this dataset are shown in Table \ref{nus-wide}, which is clearly to observe that GM-MLIC can not only effectively learn from such large-scale data, but also achieve superior performance on most evaluation metrics with 62.2\% mAP, 61.0\% CF1, 74.1\% (72.5\%) OF1 (Top-3), respectively.


\begin{table}
\begin{center}
\begin{tabular}{|c||c|c|c|c||c|}
\hline
Dataset &\multicolumn{4}{c|}{MS-COCO}  &VOC \\ \hline
Methods &mAP &CP &CR &CF1 &mAP \\\hline
Our w/o $\mathbb{G}^l$  &81.7 &85.1 &68.3 &76.5 &92.8 \\
Our w/o $\mathbb{G}^o$  &82.9 &86.4 &67.6 &77.2 &93.3 \\
Our w/o ILM edges  &83.8 &86.7 &69.2 &77.6 &94.2 \\ \hline\hline
Our GM-MLIC   &84.3 &87.3 &70.8 &78.3 &94.7 \\
\hline
\end{tabular}
\vspace{-3mm}
\caption{ Comparison of mAP (\%) of our framework (Our GM-MLIC), our framework without label semantic graph (Our w/o $\mathbb{G}^l$), our framework without object spatial graph (Our w/o $\mathbb{G}^o$) and our framework without instance-label matching edges (Our w/o ILM edges) on the VOC 2007 and MS-COCO dataset.}
\label{sub-graph}
\end{center}
\vspace{-7mm}
\end{table}

\subsection{Further Analysis}
\textbf{Ablation Studies:} To evaluate the effectiveness of each component in our proposed framework, we conduct ablation studies on the MS-COCO and VOC 2007 datasets, as shown in Table \ref{sub-graph}. It is not surprising to conclude that combining all components would achieve the best performance. Specifically, using label semantic graph $\mathbb{G}^l$ would result in 1.3\% higher precision but 0.7\% lower recall score than using instance spatial graph $\mathbb{G}^o$ solely. In other words, introducing $\mathbb{G}^o$ into our framework would reduce the possibility of missing instances in images. Besides, the instance-label matching edge that incorporating category semantics can guide the model to better learn semantic-specific features and further improve the classification precision.

\begin{figure}[t]
\vspace{-1mm}
\centering  
	\subfigcapskip=-2pt 
	\subfigure[The number of instance.]{
		\includegraphics[width=41mm,height=25mm]{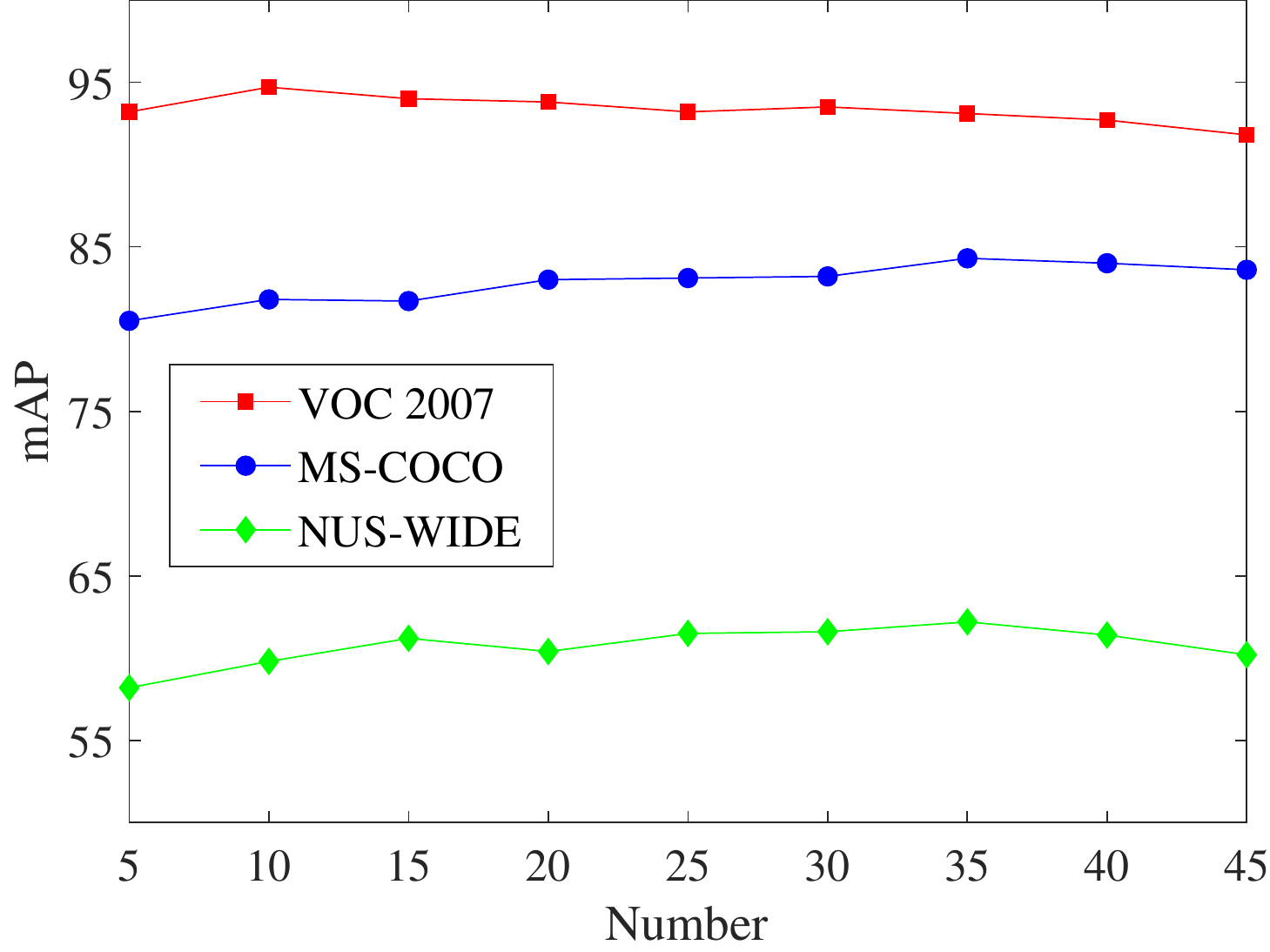}}
	\subfigure[The number of conv module.]{
		\includegraphics[width=41mm,height=25mm]{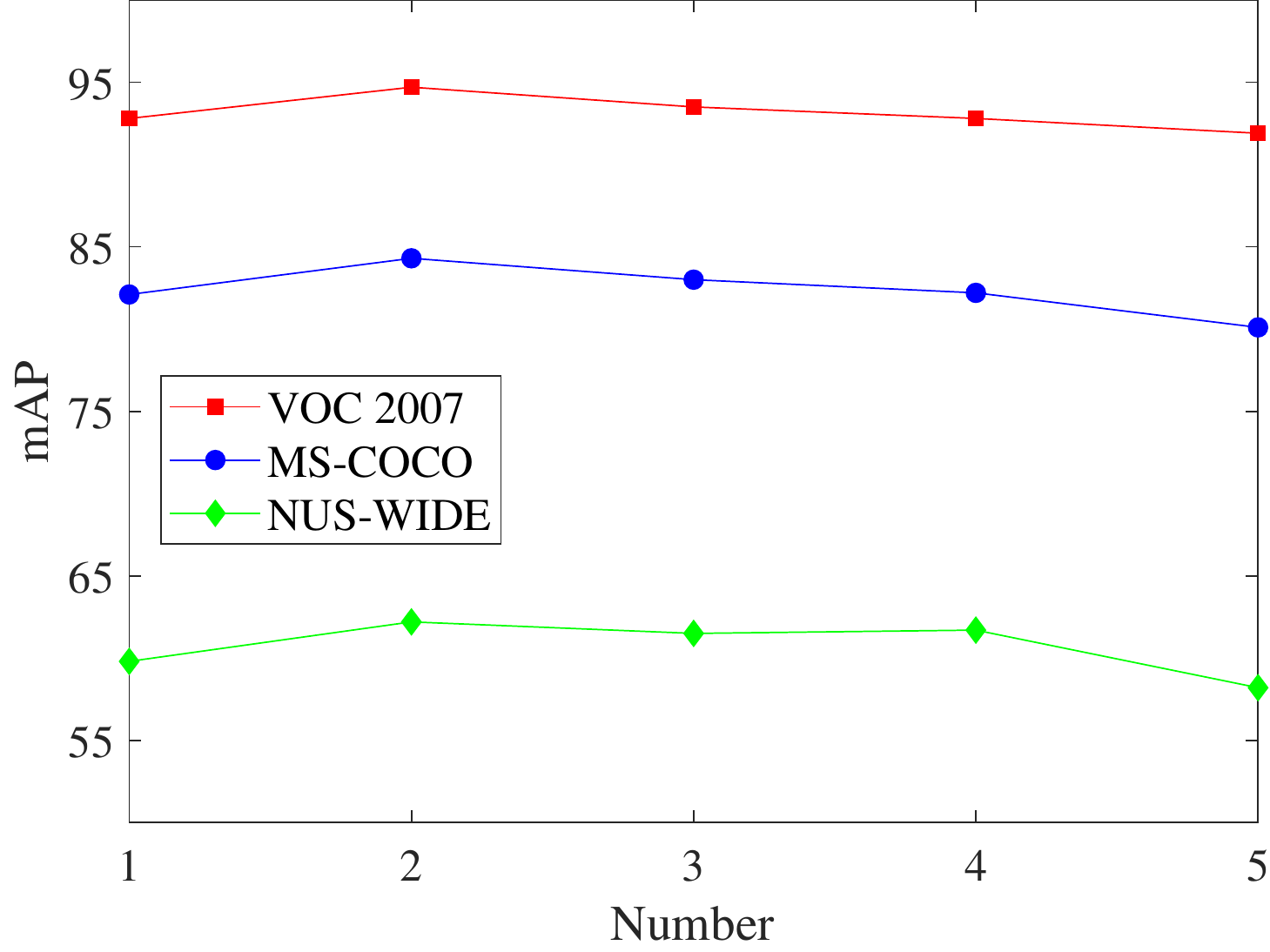}}
	\vspace{-5mm}
	\caption{Accuracy comparisons with different parameter values.}
	\label{parameter}
\vspace{-2mm}
\end{figure}

\begin{figure}[t]
\centering
\vspace{-2mm}
\subfigbottomskip=-10pt
\subfigure{
\begin{minipage}{0.5\textwidth}
\includegraphics[width=85mm,height=43mm]{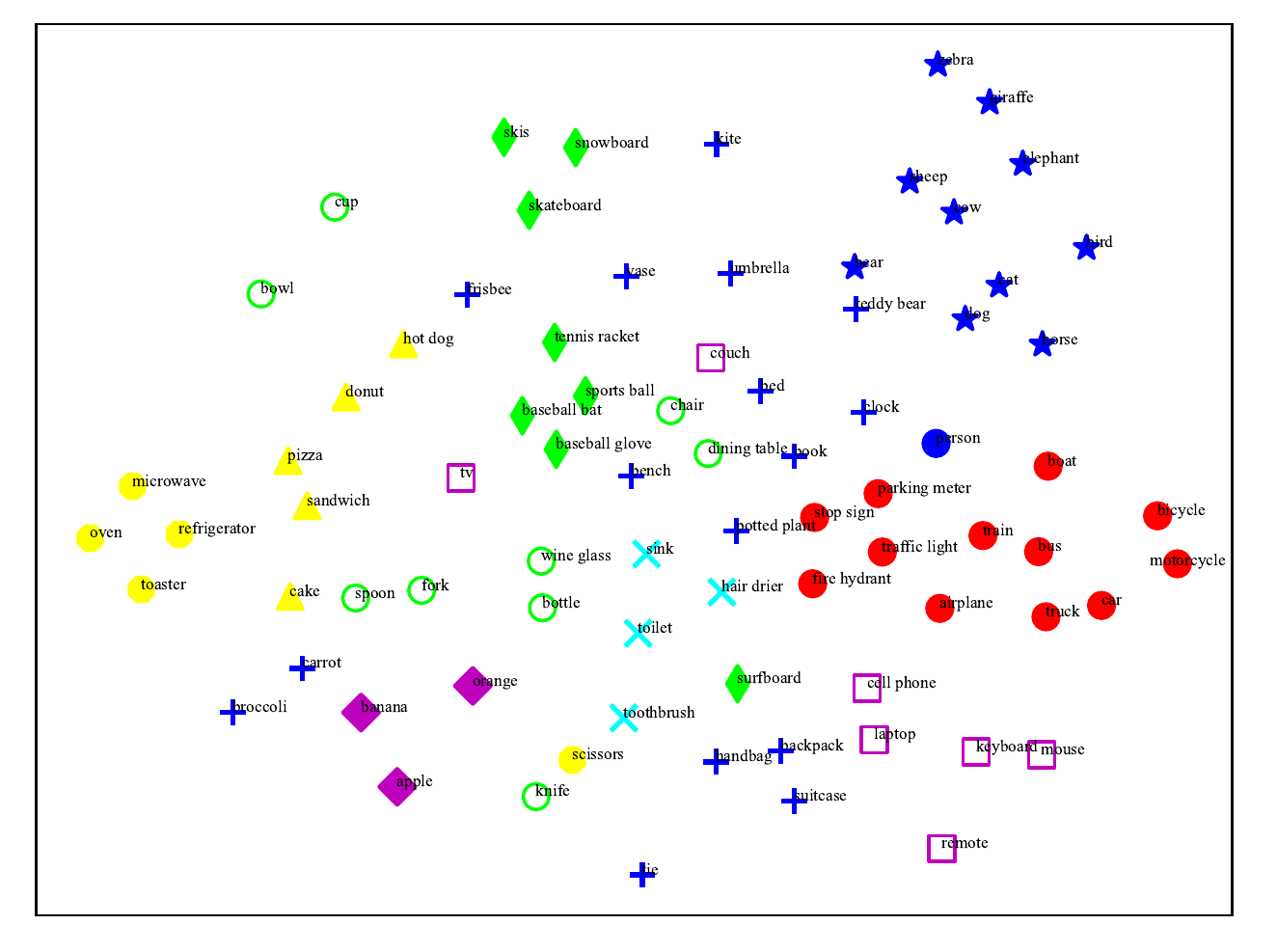}
\end{minipage}}

\subfigure{
\hspace{-0.5mm}
\begin{minipage}{0.5\textwidth}
\includegraphics[width=83mm,height=10mm]{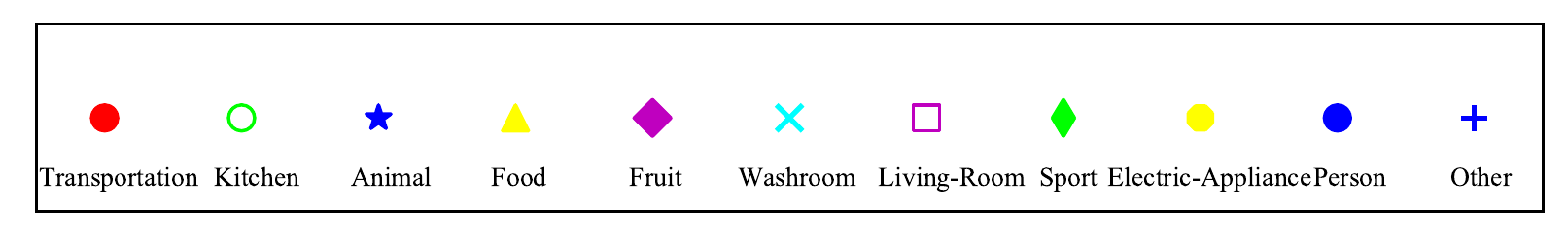}
\end{minipage}}

\caption{Visualization of the learned classifiers by our model on MS-COCO dataset.}
\label{label_cor}
 \vspace{-3mm}
\end{figure}

\textbf{Hyper-parameter and Visualization:} Furthermore, we study the performance of our proposed method given different parameter settings. We first vary the number $m$ of the extracted instances per image, and show the results in Figure \ref{parameter} (a). Note that, if we keep all instances, the model will contain a lot of noise instances, which is difficult to converge. However, when too many instances are filtered out, the accuracy drops since some positive instances in images may be removed incorrectly. Empirically, the optimal value of $m$ is set to 10 in VOC 2007 dataset, and 35 in both MS-COCO and NUS-WIDE datasets. Then, we show the performance results with different numbers of convolution modules $k$ for our model in Figure \ref{parameter} (b). With the number of convolution modules increases, the accuracy drops on all the three datasets. Thus, we empirically set $k=2$ in our experiments. In Figure \ref{label_cor}, we visualize the classifiers learned by our proposed method, which demonstrates that the meaningful semantic topology is maintained.

\section{Conclusion}
In this paper, we propose a novel graph matching based multi-label image classification learning framework GM-MLIC, which reformulates the MLIC problem into a graph matching structure. By incorporating instance spatial graph and label semantic graph, and establishing instance-label assignment, the proposed GM-MLIC utilizes graph network block to form structured representations for each instance and label by convolving its neighborhoods, which can effectively contribute the label dependencies and semantic-aware features to the learning model. Extensive experiments on various image datasets demonstrate the superiority of our proposed method.
 \vspace{-5mm}
\section*{Acknowledgments}
This work was supported by the National Natural Science Foundation of China (Nos. 61872032, 61972030, 62072027, 62076021), the Beijing Natural Science Foundation (Nos. 4202058, 4202057, 4202060), and in part by the Fundamental Research Funds for the Central universities (Nos. 2020YJS036, 2019YJS044, 2020YJS026).

\bibliographystyle{named}
\bibliography{ijcai21}

\end{document}